\documentclass[conference]{IEEEtran}

\usepackage{graphicx,times,amsmath,cite} 

\IEEEoverridecommandlockouts	
				
\newcommand{\commentaire}[1]{}
\begin{document}
\title{On the influence of selection operators on performances in cellular Genetic Algorithms}
            
\author{D. Simoncini, P. Collard, S. Verel, M. Clergue}

\maketitle

\begin{abstract}

In this paper, we study the influence of the selective pressure on the performance of cellular genetic algorithms. Cellular genetic algorithms are genetic algorithms where the population is embedded on a toroidal grid. This structure makes the propagation of the best so far individual slow down, and allows to keep in the population potentially good solutions. We present two selective pressure reducing strategies in order to slow down even more the best solution propagation.
We experiment these strategies on a hard optimization problem, the Quadratic Assignment Problem, and we show that there is a threshold value of the control parameter for both which gives the best performance. This optimal value does not find explanation on the selective pressure only, measured either by takeover time or diversity evolution. This study makes us conclude that we need other tools than the sole selective pressure measures to explain the performance of cellular genetic algorithms.
 
\end{abstract}

\section*{Introduction}

The selective pressure can be seen as the ability for solutions to survive 
in the population. When the selective pressure is high, only the best 
solutions survive and colonize the population, allowing less time for 
the algorithm to explore the search space. Thus, 
the selective pressure has an impact on the exploration/exploitation trade-off:
When it is too low, good solutions' influence on the population is so weak 
that the algorithm can't converge and behave as a random search in the 
search space. When it is too strong, the algorithm converges quickly 
and as soon as it is stuck in a local optimum it won't be able to 
find better solutions.

Cellular Genetic Algorithms (cGA) are a subclass of Evolutionary Algorithms in 
which the population is embedded on a bidimensional toroidal grid.
Each cell of the grid contains one individual (solution) and the stochastic 
operators are applied within the neighborhoods of each cell.
The existence of such small overlapped neighborhoods guarantee the propagation
 of solutions through the grid and enhance exploration and population 
diversity \cite{SpiessensM91}. Such a kind of algorithms is especially well 
suited for complex problems with multiple local optima \cite{JongS95}.
To avoid the algorithm to converge toward one local optimum, one should 
apply the right selective pressure on the population and find the 
best balance between exploitation of good solutions and exploration of the 
search space.

Section 1 presents a state of the art on selective pressure in cGAs and 
introduces two selection operators. Section 2 compares the influence of 
the selection operators on the selective pressure. Section 3 gives a description 
of the benchmark used to analyze the algorithms. Section 4 presents a comparative 
study of performance of the algorithms. Section 5 is a study on the evolution 
of the genotypic diversity in the populations. Finally in section 6 we summarize 
and discuss the results of the paper.

\section{Cellular genetic algorithms and selective pressure}
\label{section2}

Several methods have been proposed to tune the selective pressure and 
deal with the exploration/exploitation trade-off in cGA. For instance, 
the size and shape of the cells neighborhoods in which the evolutionary 
operators are applied, has some influence. A bigger neighborhood will induce 
a stronger selective pressure on the population \cite{SarmaJ96}. When 
trying to solve complex problems, with numerous local optima, one would 
try to slow down the convergence of the population. That is why we use 
in our algorithm a Von Neumann neighborhood which is the smallest symetric 
neighborhood that allows the convergence of the population.

The shape of the grid also has an impact on the selective pressure \cite{AlbaT00,Giacobini2003,Giacobini2004}: 
thinner grids give a weaker selective pressure on the population.
This solution's weakness is that there are not enough grid shapes 
for a fixed size of population to allow an accurate control of 
the selective pressure.

The selective pressure can also be monitored by choosing an adequate selection 
operator.

\subsection{Stochastic tournament selection}

The stochastic tournament selection proposed by Goldberg
 is a binary tournament selection that 
doesn't guarantee the best solution to be selected. 
The stochastic tournament of rate $r$ chooses two solutions from the 
neighborhood of a cell and selects the best one with probability $1-r$
(the worst one with probability $r$). Real parameter $r$ should be 
in $[0;1]$.

Given the definition of selective pressure, this selection operator 
explicitely gives a weaker selective pressure for increasing $r$ values. 
As $r$ is getting closer to $1$, worse solutions increase their 
chances to be maintained in the population, which means the selective pressure
is getting weaker.

\subsection{Anisotropic selection}

The Anisotropic selection is a selection method in which the neighbors of a cell may have different probabilities to be selected \cite{gecco06}.
The Von Neumann neighborhood of a cell $C$ is defined as the sphere of radius $1$ centered at $C$ in manhattan distance. 
 The Anisotropic selection assigns different 
 probabilities to be selected to the cells of the Von Neumann neighborhood according to their position. The probability $p_c$ to choose the center cell $C$ 
remains fixed at $\frac15$. Let us call $p_{ns}$ the probability of choosing the cells North ($N$) or South ($S$) and $p_{ew}$ the probability of choosing the cells East  
($E$) or West ($W$). Let $\alpha \in [-1;1]$ be the control parameter that will determine the probabilities $p_{ns}$ and $p_{ew}$. This parameter will be called 
the \textit{anisotropic degree}.
The probabilities $p_{ns}$ and $p_{ew}$ can be described as: \\
$$p_{ns}=\frac{(1-p_c)}{2}(1+\alpha)$$
$$p_{ew}=\frac{(1-p_c)}{2}(1-\alpha)$$ 
Thus, when $\alpha=-1$ we have $p_{ew}=1-p_c$ and $p_{ns}=0$. When $\alpha=0$, we have $p_{ns}=p_{ew}$ and when $\alpha=1$, we have $p_{ns}=1-p_c$ and $p_{ew}=0$.
In the following, the probability $p_c$ remains fixed at $\frac{1}{5}$.

\begin{figure}[ht!]
\begin{center}
\includegraphics[width=3cm,height=3cm]{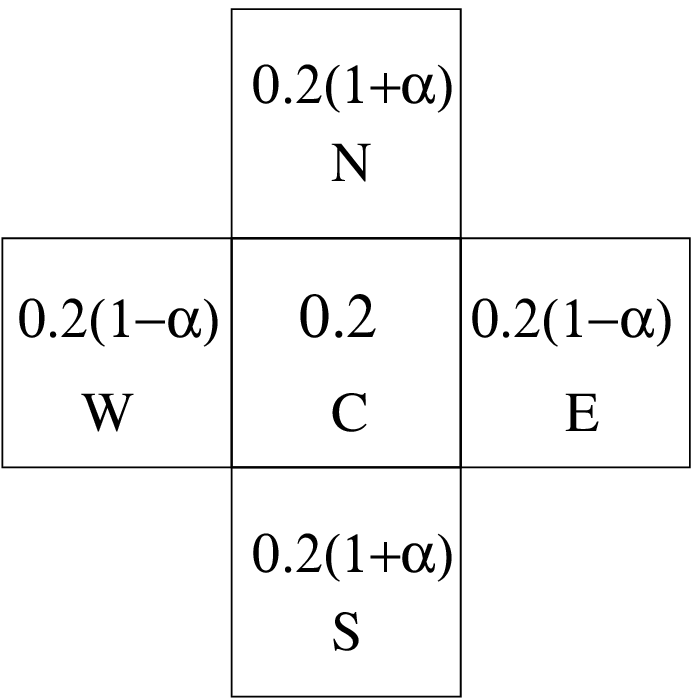} 
\caption{Von Neumann neighborhood with probabilities to choose each neighbor}
\end{center}
\label{VN}
\end{figure}

Figure $1$ shows a Von Neumann Neighborhood with the probabilities to select each cell as a function of 
$\alpha$.

The Anisotropic Selection operator works as follows. 
For each cell it selects $k$ individuals in its neighborhood ($k \in [1;5]$). The $k$ individuals participate to a tournament and the winner replaces 
the old individual if it has a better fitness or with probability $0.5$ if the fitnesses are equal. 
When $\alpha=0$, the anisotropic selection is equivalent to a standard tournament selection and when $\alpha=1$ or $\alpha=-1$ the anisotropy is maximal and we 
have an uni-dimensional neighborhood with three neighbors only. In the following, considering the grid symmetry, we will consider $\alpha \in [0;1]$ only: when $\alpha$ is in the 
range [-1;0] making a rotation of $90^{\circ}$ of the grid is equivalent to 
considering $\alpha$ in the range [0;1].

\section{Takeover time}

A common analytical approach to measure the selective pressure is the computation of the takeover time \cite{Rudolph} \cite{Sprave}. It is the time 
needed for the best solution to colonize the whole population when 
the only active evolutionary operator is selection \cite{GoldbergD90}. When the takeover time 
is short, it means that the best solution's propagation speed in the 
population is high. So, worse solutions' life time in the population 
is short and thus the selective pressure is strong.
 On the other hand, when the 
takeover time is high, it means that the best solution colonizes slowly 
the population, giving a longer lifetime to worse solutions. 
In that case, the selective pressure is low. 
So the selective pressure in the population is inversely proportionnal to the takeover time. 

In order to measure the takeover time, we place one solution of fitness $1$ on a $20\times 20$
grid. All the other solutions have a null fitness. Then we run the process and measure 
the time needed for the solution of fitness $1$ to spread over the whole grid. 
 
\begin{figure}
\begin{center}
\begin{tabular}{c}
\includegraphics[width=6cm,height=6cm]{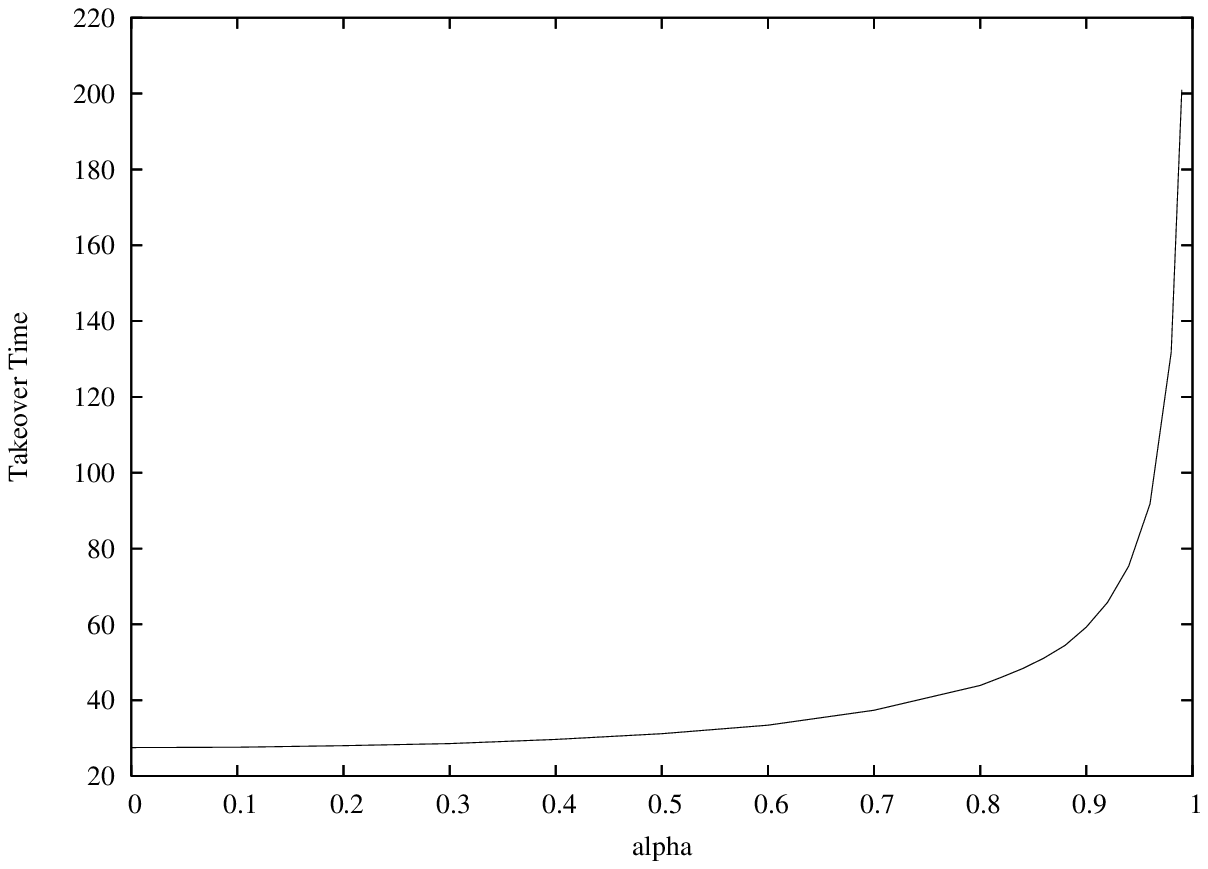} \\
(a) \\
\includegraphics[width=6cm,height=6cm]{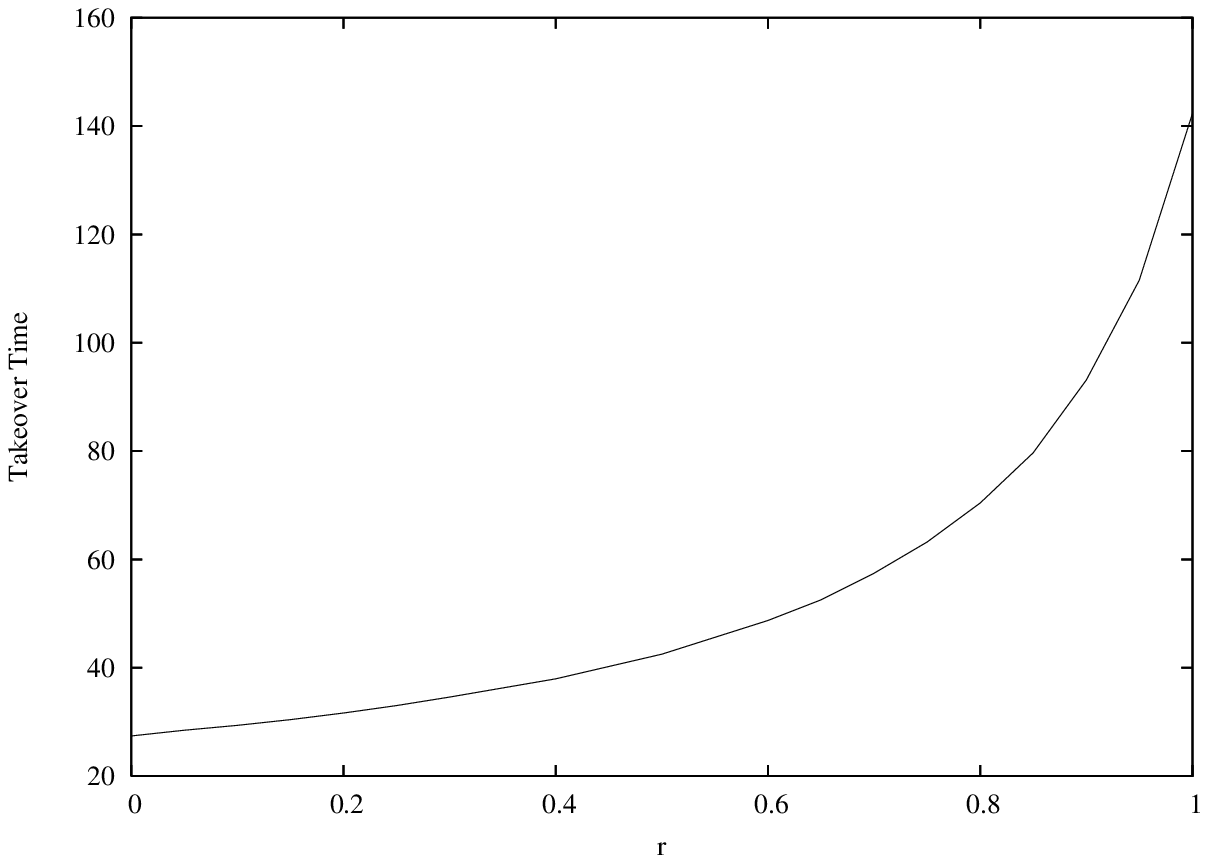} \\
(b) \\
\end{tabular}
\end{center}
\caption{Average takeover times for a cGA using anisotropic selection (a) 
and stochastic tournament selection (b)}
\label{takeover-time}
\end{figure}

We measured average takeover times over $1000$ simulations for a cGA using 
 a stochastic tournament selection, and for one using the anisotropic 
selection. The simulations are made on square grids of side $20$.
Figure \ref{takeover-time} shows the results of these simulations.
The takeover time increases when $\alpha$ increases in the 
case of a cGA using the anisotropic selection (figure \ref{takeover-time}(a)).
So the selective pressure is inversely proportional to $\alpha$. 
On figure \ref{takeover-time}(b) we can see that the takeover time increases
 as long as the probability $r$ to select the worst solution in the 
stochastic tournament grows. This means that the selective pressure in the 
population is inversely proportional to $r$ for a cGA using  a stochastic 
tournament selection. 

The slope of the curve representing the takeover time as a function of $\alpha$ (fig. \ref{takeover-time}(a)) for values close to $1$ is more important than the one of the curve representing the takeover time 
as a function of $r$ (fig. \ref{takeover-time}(b)).   
We can also notice that in the case of a cGA using stochastic tournament selection, the takeover time is defined 
when the probability to select the best solution is $0$. The best solution still can colonize the population in this case since the two candidates for the tournament are selected by a random draw with replacement. 
In the case of a cGA using anisotropic selection, the takeover time is not defined for $\alpha=1$.The anisotropic degree $\alpha$ is a continuous parameter and the curve representing 
the takeover time as a function of $\alpha$ is not bounded.  

\section{The Quadratic Assignment Problem}

This section presents the Quadratic Assignment Problem (QAP) which is known to be difficult to optimize.
The QAP is an important problem in theory and practice as well. It was introduced by Koopmans and Beckmann
in 1957 and is a model for many practical problems \cite{Koopmans57}. 
The QAP can be described as the problem of assigning a set of facilities to
a set of locations with given distances between the locations and given flows between the 
facilities. The goal is to place the facilities on locations in such a way that the sum
 of the products between flows and distances is minimal.
\linebreak
Given $n$ facilities and $n$ locations, two $n \times n$ matrices $D=[d_{ij}]$ and $F=[f_{kl}]$
 where $d_{ij}$ is the distance between locations $i$ and $j$ and $f_{kl}$ the flow between 
 facilities $k$ and $l$, the objective function is: \\
\begin{displaymath}
\Phi = \sum_{i}\sum_{j}d_{p(i)p(j)}f_{ij}
\end{displaymath}
where $p(i)$ gives the location of facility $i$ in the current permutation $p$.
\linebreak
 Nugent, Vollman and Ruml proposed a set of problem instances of different sizes noted for their difficulty \cite{Nugent68}. The instances they proposed are known to have multiple local optima, so they are difficult for a genetic algorithm. We experiment our algorithm on the instances nug30 (30 variables), 
tho40 (40 variables) and sko49 (49 variables) from QAPLIB.

\subsection*{Set up}

We use a population of 400 individuals placed on a square grid ($20\times 20$). Each individual
 is reprensented by
 a permutation of $N$ where $N$ is the size of an individual.
 The algorithm uses a crossover that preserves the permutations:
\begin{itemize}
\item
  Select two individuals $p_1$ and $p_2$ as genitors.
\item
  Choose a random position $i$.
\item
  Find $j$ and $k$ so that $p_1(i) = p_2(j)$ and $p_2(i) = p_1(k)$.
\item
  exchange positions $i$ and $j$ from $p_1$ and positions $i$ and $k$ from $p_2$.
\item
  repeat $N/3$ times this procedure where $N$ is the size of an individual.
\end{itemize}

This crossover is an extended version of the UPMX crossover proposed in \cite{Migkikh}.
The mutation operator consist in randomly selecting two positions from the individual
 and exchanging those positions. The crossover rate is 1 and we do a mutation per individual.
We perform 200 runs for each tuning of the two selection operators. An elitism replacement 
procedure guarantees the individuals to stay on the grid if they are fitter than their 
offspring.
 Each run stops after 2000 generations for nug30 and tho40, and after 
3000 generations for sko49.

\section{Performances}

In this section we present performance results on the Quadratic Assignment 
Problem for a cGA using stochastic tournament and anisotropic selection operators.
In \cite{acri06} the authors show that there is an optimal value of $\alpha$ parameter 
for the anisotropic selection that gives optimal performance. 
We want to see if the same behaviour is observed with 
the stochastic tournament selection and then to compare the 
performance obtained for these two operators.

\begin{figure}[h!]
\begin{center}
\begin{tabular}{c}
\includegraphics[width=6cm,height=6cm]{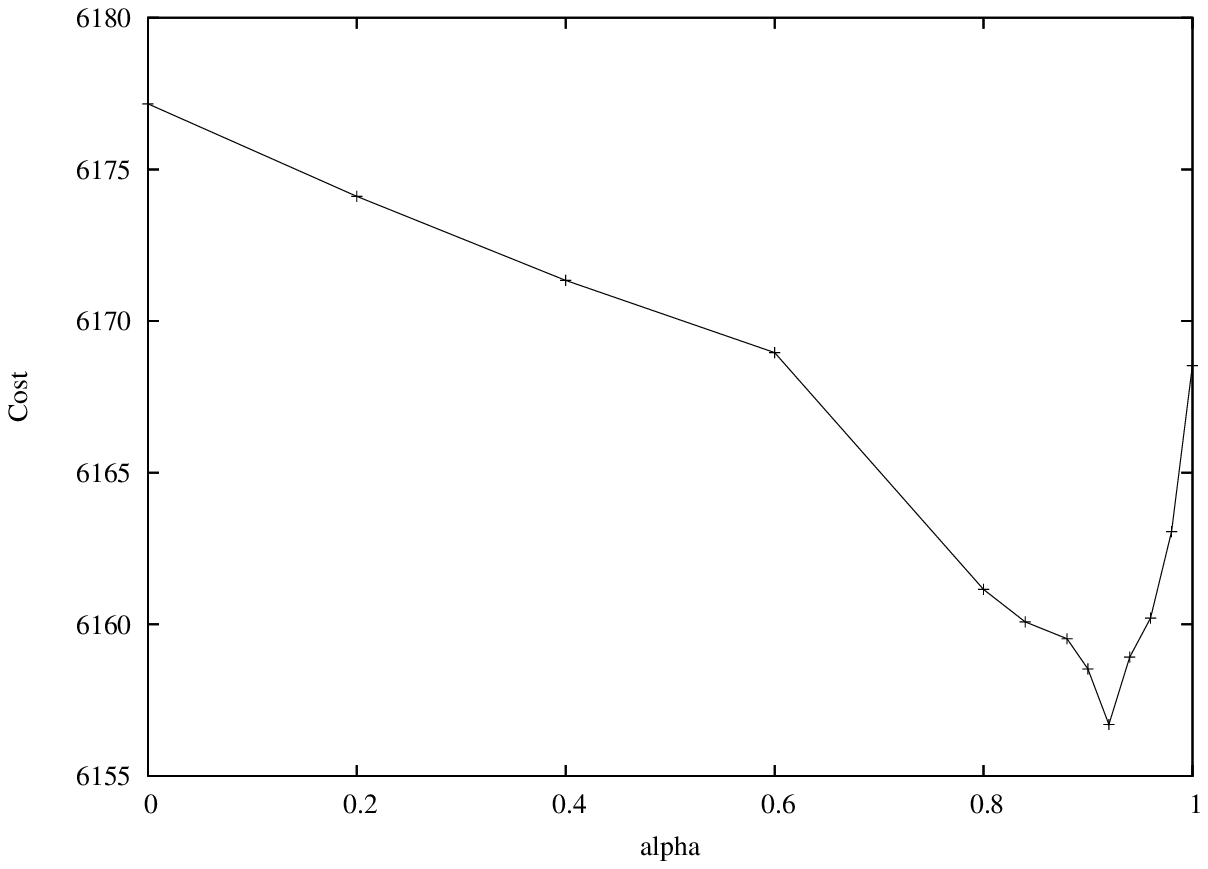} \\
(a) \\
\includegraphics[width=6cm,height=6cm]{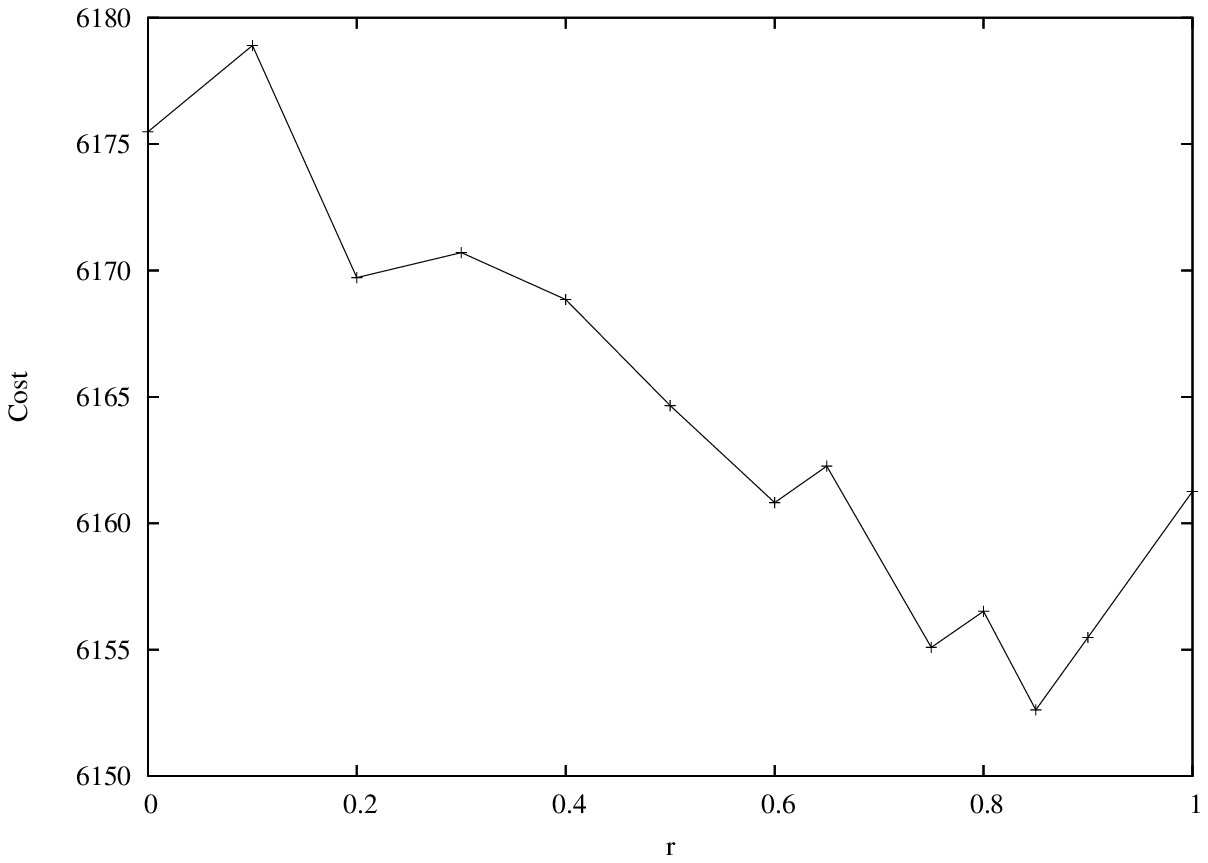} \\
(b) \\
\end{tabular}
\end{center}
\caption{Performance of cGA with anisotropic selection for different anisotropy degrees (a) and with stochastic tournament for different rates (b) on instance nug30}
\label{performances}
\end{figure}

\begin{figure}[h!]
\begin{center}
\begin{tabular}{c}
\includegraphics[width=6cm,height=6cm]{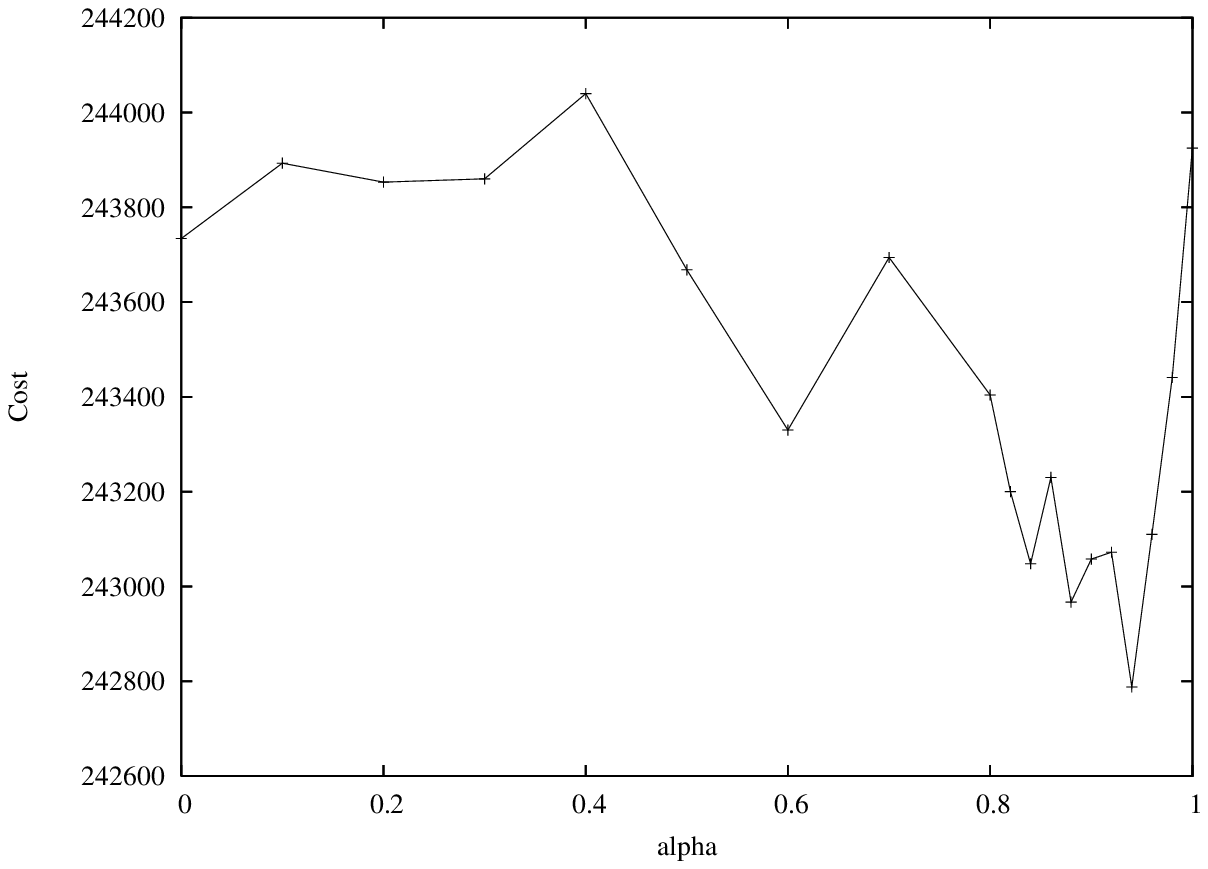} \\
(a) \\
\includegraphics[width=6cm,height=6cm]{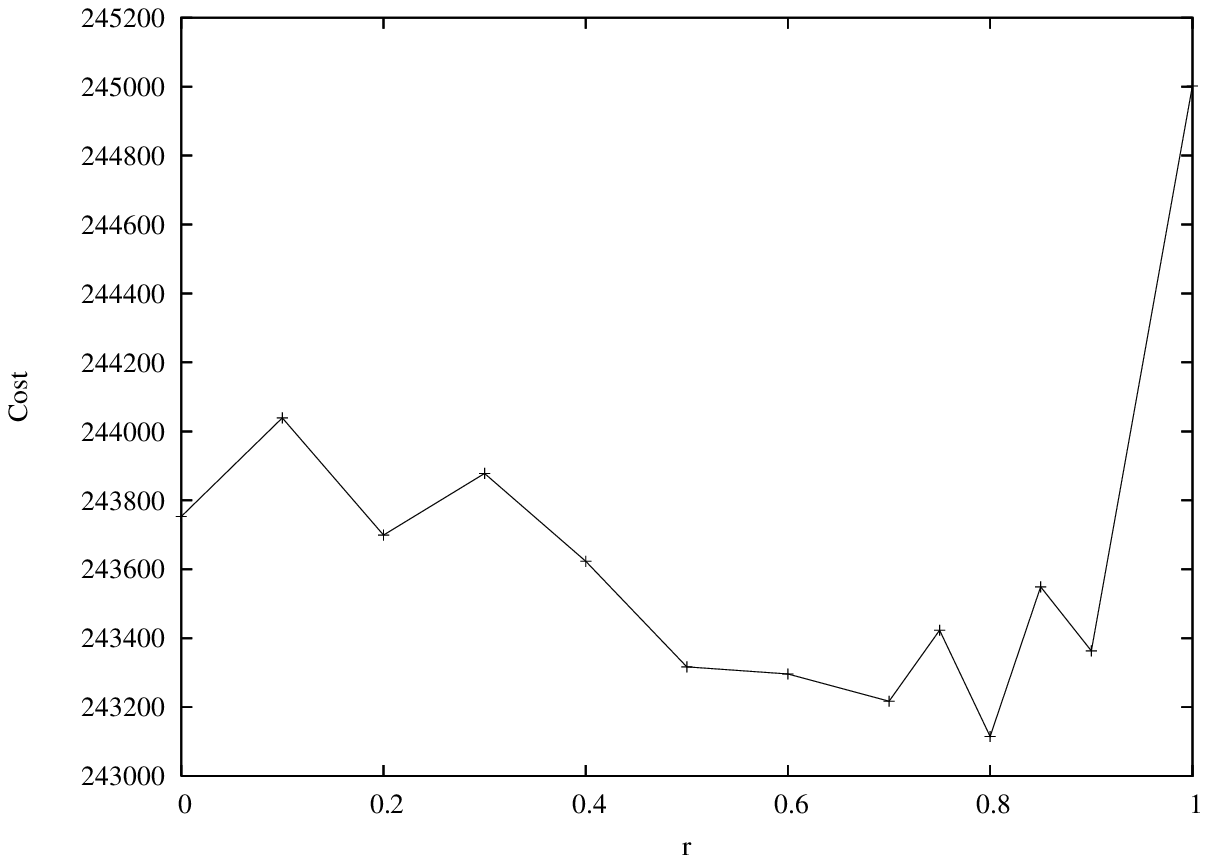} \\
(b) \\
\end{tabular}
\end{center}
\caption{Performance of cGA with anisotropic selection for different anisotropy degrees (a) and with stochastic tournament for different rates (b) on instance tho40}
\label{performances-tho}
\end{figure}

\begin{figure}[h!]
\begin{center}
\begin{tabular}{c}
\includegraphics[width=6cm,height=6cm]{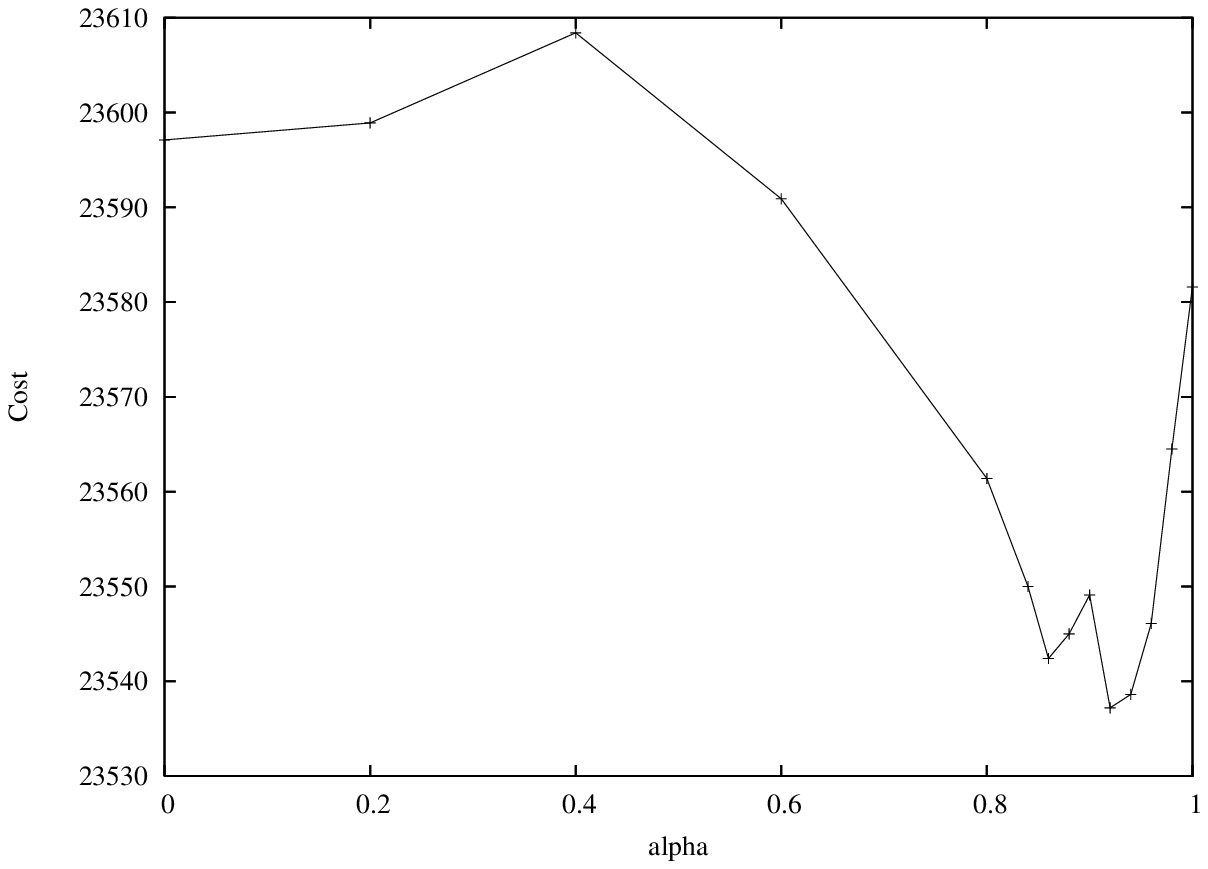} \\
(a) \\
\includegraphics[width=6cm,height=6cm]{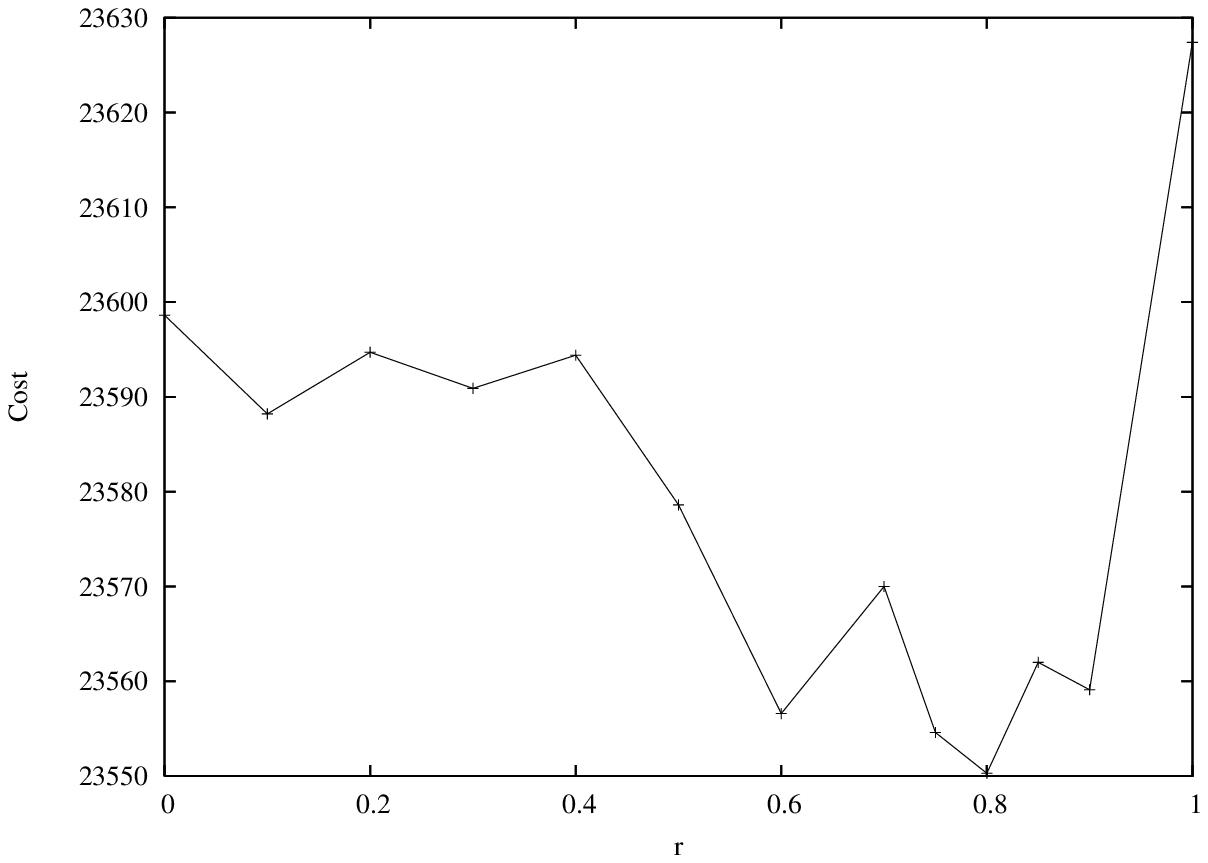} \\
(b) \\
\end{tabular}
\end{center}
\caption{Performance of cGA with anisotropic selection for different anisotropy degrees (a) and with stochastic tournament for different rates (b) on instance sko49}
\label{performances-sko}
\end{figure}

Figures \ref{performances}, \ref{performances-tho} and \ref{performances-sko}
  show performance obtained with 
the anisotropic and the stochastic selection operator on the QAP instances 
nug30, tho40 and sko49. 
We measure the performance by averaging the best solution found on 
each run for each value of anisotropy degree and stochastic rate.

When the rate of the stochastic tournament selection and the anisotropic 
degree are null, 
the two algorithms are the same : A standard cGA with binary tournament 
selection. 
The selective pressure drops when the values of the control parameters of the 
two algorithms increase.
In both cases we see that as the selective pressure drops, performance 
increases until a threshold value. Once this value is reached, the performance
decreases. 
These threshold values give the best exploration/exploitation trade-off for this problem. 
In the following, the threshold values of parameters $\alpha$ and $r$ are 
denoted $\alpha_o$ and $r_o$.

\begin{table}
\begin{center}
\caption{Average performance and Takeover times for $\alpha_o$ and $r_o$}
\begin{tabular}{|r|r r r|r r r|}
\hline
         & \multicolumn{3}{|c|}{Anisotropic selection} &
           \multicolumn{3}{|c|}{Stochastic tournament selection}\\
Instance & Perf & Val & TO & Perf & Val & TO \\
\hline
Nug30 & $6156.3_{18.6}$ & $0.92$ & $65.7$ & \textbf{6152.6}$_{18.5}$ & $0.85$ & $79.6$ \\ 
Tho40 & \textbf{242788}$_{988.4}$ & $0.94$ & $75.3$ & $243115_{1177.2}$ & $0.8$ & $70.4$ \\ 
Sko49 & \textbf{23537.2}$_{55.5}$ & $0.92$ & $65.7$ & $23550.3_{58}$ & $0.8$ & $70.4$ \\
\hline
\end{tabular} 
\end{center}
\label{tab-takeover-rec}
\end{table}

Table 1 gives $\alpha_o$ and $r_o$ for each instance of QAP and their corresponding takeover times (TO). Best performances are in bold and differences between performances of the two algorithms are statistically significant for each instance according to the Student's t-test. Differences in takeover times are also statistically significant. 
The algorithm using stochastic tournament selection is the best for nug30, and 
the one using anisotropic selection is the best for tho40 and sko49.
The threshold values stand in the same ranges for all instances: $\alpha_o \in \lbrace 0.92,0.94
 \rbrace$ 
and $r_o \in \lbrace 0.8,0.85 \rbrace$. 
The differences in takeover times indicate that the selective pressure on 
the population is different for the two methods for the settings that give 
the best average performance. 
These differences can be explained by the way the algorithm explores the 
search space and exploits good solutions.

\section{Diversity}

In this section, we present statistic measures on the evolution of the genotypic diversity in the population. 
Three kinds of measures are performed : The global average diversity, the 
vertical/horizontal diversity and the local diversity. 
The global average diversity measure is made on a set of $50$ runs of one instance of QAP for each kind of algorithm.
It consists in computing the genotypic diversity between each solutions generation after generation. 
$$
gD=(\frac{1}{\sharp r \sharp c})^2\sum_{r_1,r2}\sum_{c_1,c_2} d(x_{r_1c_1},x_{r_2c_2})
$$ 
where $d(x_1,x_2)$ is the distance between solutions $x_1$ and $x_2$.
The distance used is inspired from the Hamming distance: It is the number of locations that
differ between two solutions divided by their length $n$.

The results for each generation are averaged on $50$ runs. We obtain a curve representing the evolution of 
the global diversity in the population through $2000$ generations. 

The vertical/horizontal diversity measures the average diversity in the columns and 
in the rows of the grid. The vertical (resp. horizontal) diversity is the sum
of the average distance between all solutions in the same column (resp. row) divided by the 
number of columns (resp. rows):

$$
vD=\frac{1}{\sharp r}\frac{1}{\sharp c^2} \sum_{r}\sum_{c_1,c_2} d(x_{rc_1},x_{rc_2})
$$ 

$$
hD=\frac{1}{\sharp c}\frac{1}{\sharp r^2}\sum_{c}\sum_{r_1,r_2} d(x_{r_1c},x_{r_2c})
$$

\begin{figure}[h!]
\begin{center}
\begin{tabular}{c}
\includegraphics[width=6cm,height=5.8cm]{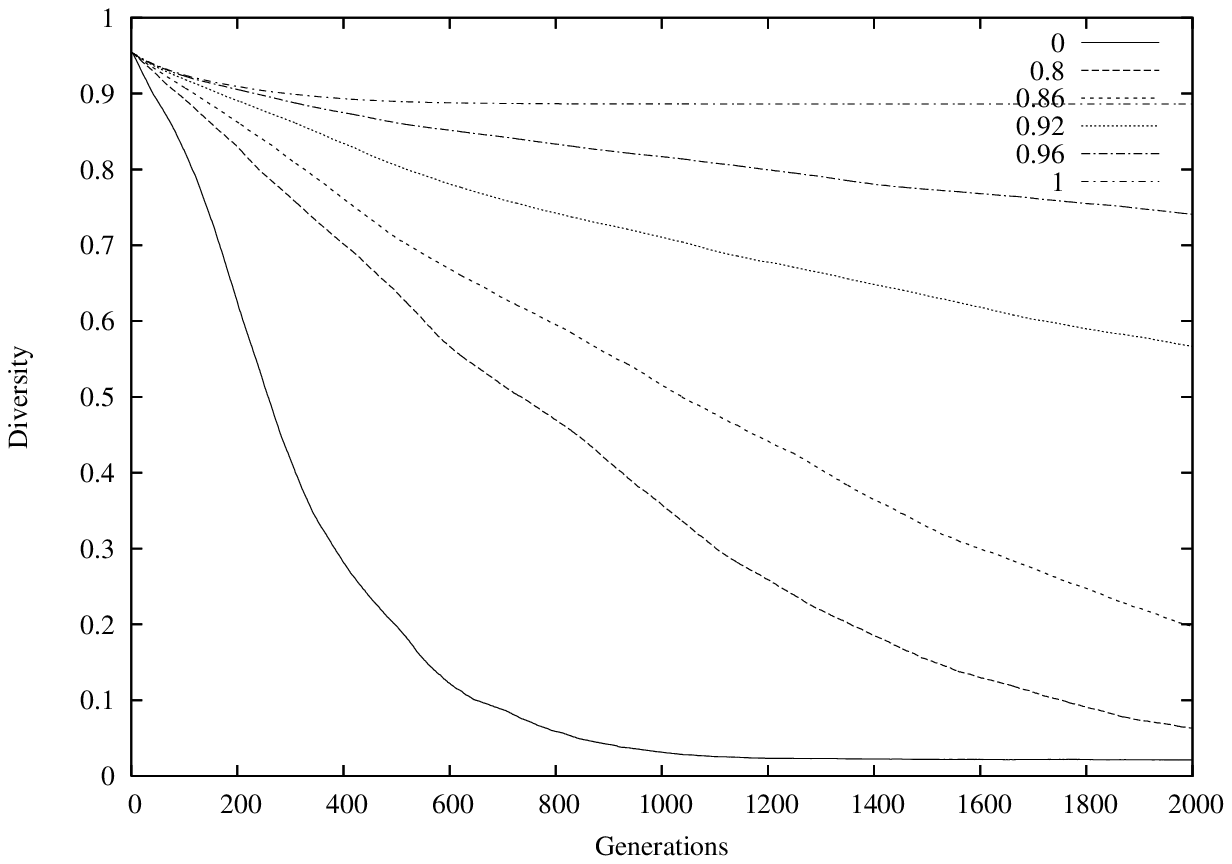} \\
(a) \\
\includegraphics[width=6cm,height=5.8cm]{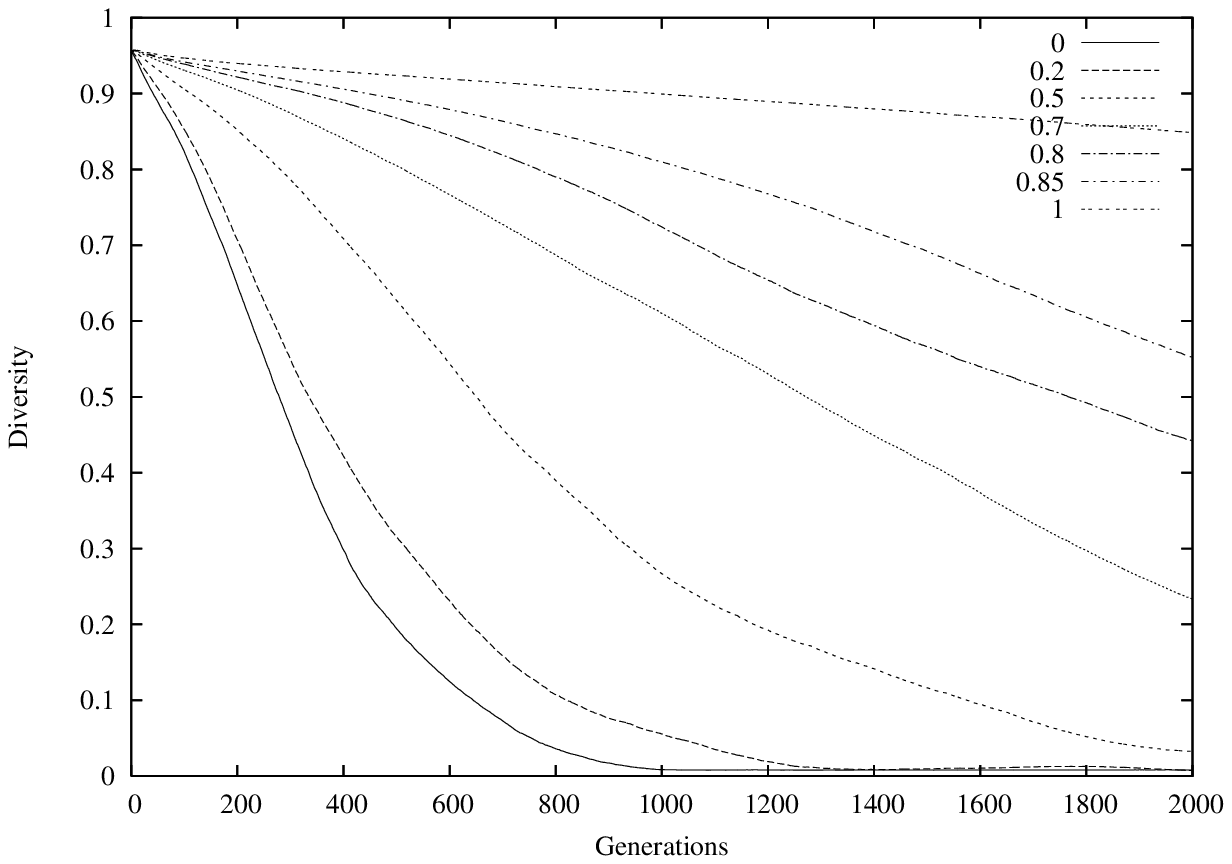} \\
(b) \\
\end{tabular}
\end{center}
\caption{Evolution of global diversity for a cGA using anisotropic selection (a) and 
stochastic tournament seletion (b) on instance nug30}
\label{divglob}
\end{figure}

\begin{figure}[h!]
\begin{center}
\includegraphics[width=6cm,height=5.8cm]{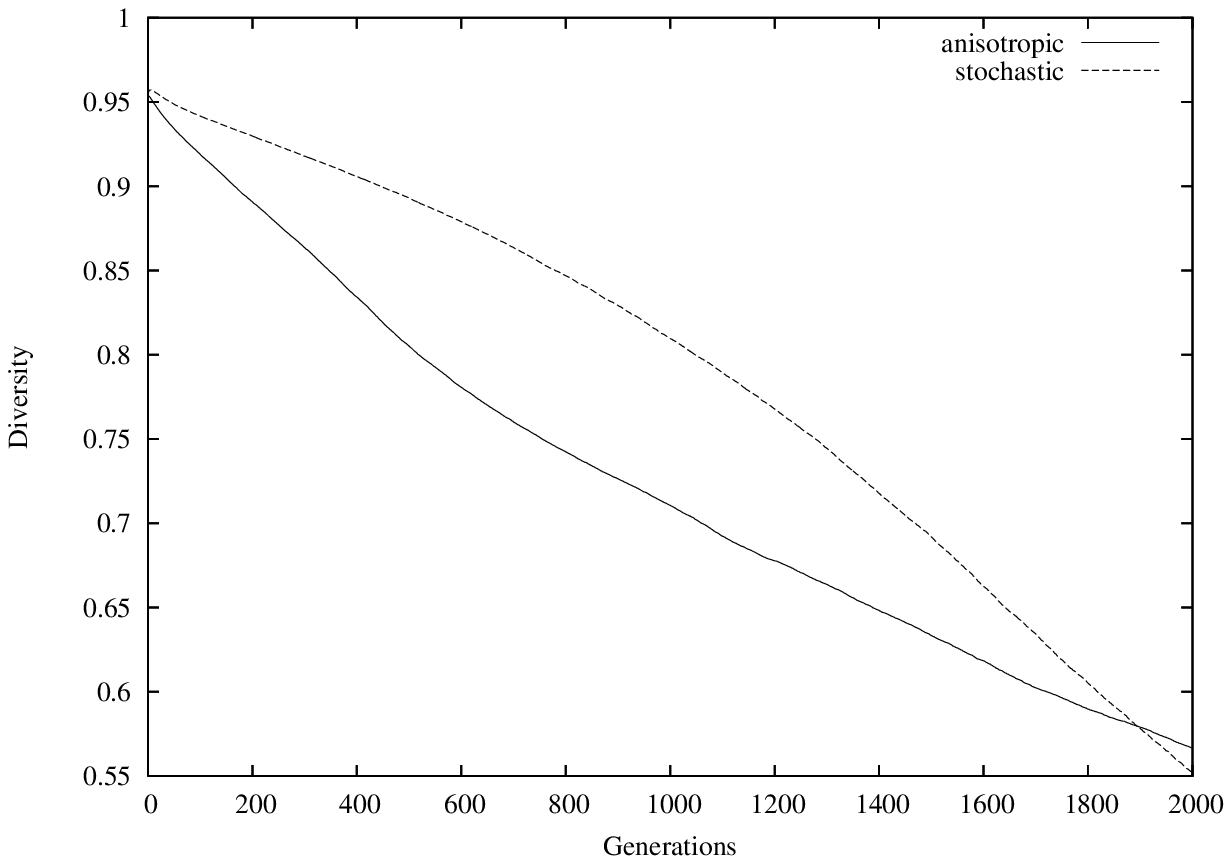} 
\end{center}
\caption{Evolution of global diversity for $\alpha_o$ and $r_o$
on instance nug30}
\label{opt-divglob}
\end{figure}

\begin{figure}[h!]
\begin{center}
  \includegraphics[width=6cm,height=5.8cm]{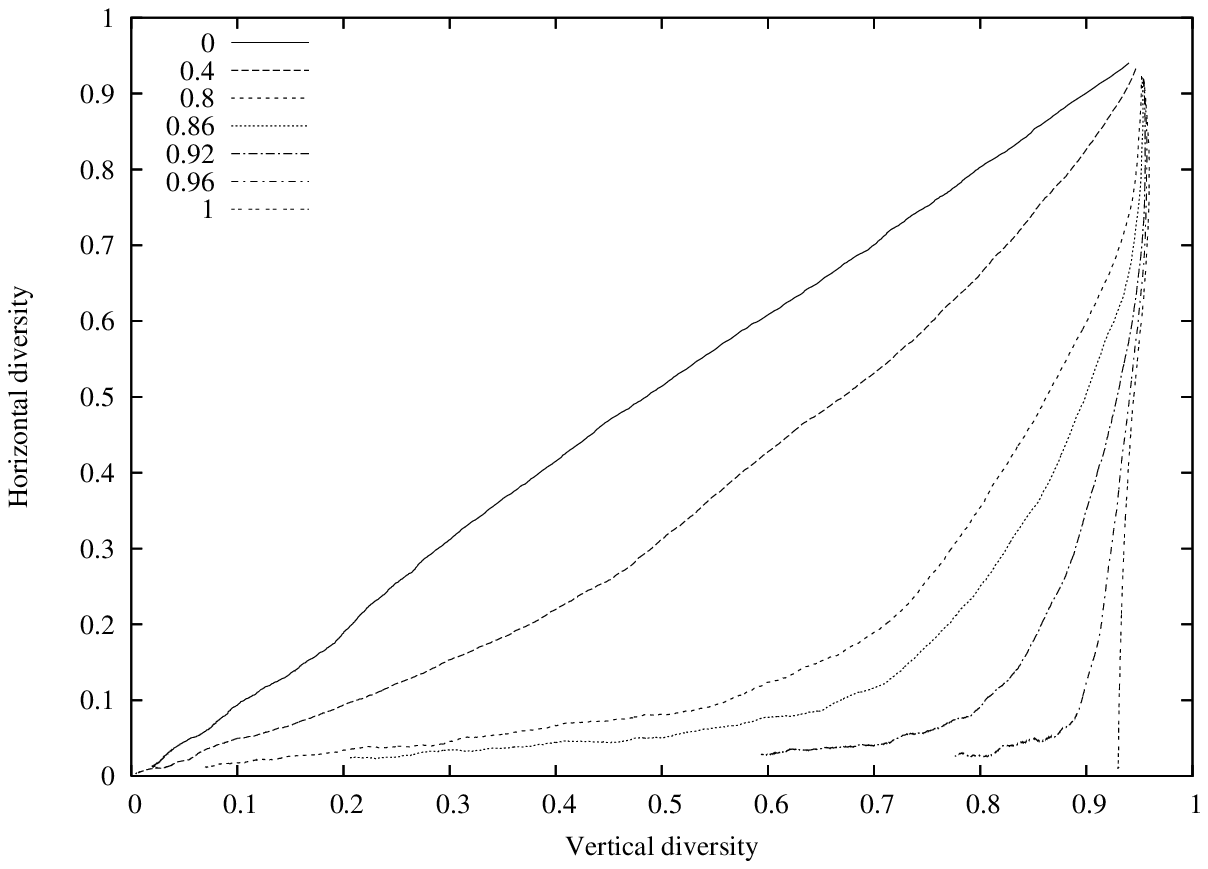} 
\end{center}
\caption{Horizontal diversity as a function of vertical diversity for different 
settings of a cGA using anisotropic selection on instance nug30}
\label{divvh}
\end{figure}

where $\sharp r$ and $\sharp c$ are the number of rows and columns in the grid.

This measure is only made for the cGA with anisotropic selection. As the stochastic tournament 
selection provides an isotropic diffusion of solutions, the difference between 
horizontal and vertical diversities is null.

Figure \ref{divglob} shows the evolution of global diversity for different settings 
of the anisotropic selection (fig \ref{divglob}(a)) and the stochastic tournament 
selection (fig \ref{divglob}(b)). 
Curves on figure \ref{divglob}(a) represent diversity for increasing values of 
$\alpha$ from bottom to top. These curves show that the more $\alpha$ is high, the more 
the diversity is maintained in the population. Similar results are obtained in the 
case of the stochastic tournament on figure \ref{divglob}(b). These curves represent 
diversity for increasing values of $r$ from bottom to top. 
The shape of the curves are different for the two methods: For the stochastic 
tournament, the curves are concave in a first time and then become convex.
For high values of $r$, the concave phase is longer and it is not finished 
at generation $2000$ for values above $0.8$. 
For the anisotropic selection, the convexity does not change, the curves are convex. 

The differences in the evolution of genotypic diversity are shown on figure \ref{opt-divglob}. 
This figure present the evolution of the diversity for the threshold values of the 
two algorithms: $\alpha_o$ and $r_o$.
 We can see that the diversity is higher for the algorithm using 
the stochastic tournament selection. Nevertheless, since the curve of the stochastic 
tournament is concave and the curve of the anisotropic selection convex, the difference 
of diversity starts to decrease around generation $1000$ and at generation $2000$ 
the algorithm using anisotropic selection has preserved more diversity. 

On figure \ref{divvh}, the horizontal diversity is plotted as a function of the vertical 
diversity for different settings of anisotropic selection. The straight line is the 
curve obtained for $\alpha=0$ and $\alpha$ is increasing on curves 
from the left to the right. As $\alpha$ increases, the algorithm favors the propagation
 of solutions in the columns of the grid and the vertical diversity decreases quicker. 
On the other hand the horizontal diversity decreases slower, and is constant for 
the limit case $\alpha=1$. In the latter case, there are no interactions between the 
columns of the grid and the algorithm behave as several independant algorithms executing 
in parallel. These algorithms run on grids of width $1$ and with $3$ solutions in the ``vertical'' neighborhoods.

The local diversity measure is computed on one single run for each kind of algorithm. It is the genotypic 
diversity observed in the neighborhood of each cell of the grid. It is represented as ``snapshots'' of the population,
 where a dark point represents a high degree of diversity in the neighborhood, and a clear point represents a low
degree of diversity in the neighborhood.

\begin{figure}
\begin{center}
\fbox{\begin{tabular}{ccccc}
$1$ &
$300$ &
$500$ & 
$1000$ & 
$1500$ \\
\includegraphics[width=2cm,height=2cm]{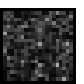} &
\includegraphics[width=2cm,height=2cm]{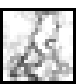} &
\includegraphics[width=2cm,height=2cm]{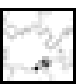} &
\includegraphics[width=2cm,height=2cm]{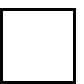} &
\includegraphics[width=2cm,height=2cm]{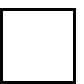} \\
\end{tabular}}
\end{center}
\caption{Local diversity in the population along generations (left to right) for a
cGA with standard tournament selection}
\label{divers-standard}
\end{figure}

Figure \ref{divers-standard} represents the local diversity along generations
for a cGA with standard binary tournament selection. We can see on snapshots 
of generations $300$ and $500$ the formation of circles. Each circle contains
 copies of good solutions found locally. The frontier between the areas from which good solutions colonize the grid are the only sites on the grid where the crossover operator still can have some effect. At generation $1000$ the genotypic diversity on the grid is null
, the population has been colonized by one solution, and performance will not improve anymore.

Figure \ref{divers-stochastic} represents the local diversity along generations
for a cGA with stochastic tournament selection. 
The probability of selecting 
the best participant to the tournament is decreasing from top to bottom.
The first thing we notice is that the propagation mode of good solutions is the
same as for the cGA using a standard binary tournament selection. The only 
difference is the speed of propagation of good solutions monitored by the $r$ 
parameter. As long as we give less chances to the best solution in the
neighborhood to be selected, it will take more time before the algorithm 
converges. The areas where crossovers can help to explore the search space 
and to improve performance are bigger when $r$ increases. The crossover 
operator does not have any effect in the white zones of the grid since 
there is no more genotypic diversity in such areas.
 For $r=1$, we see that at generation $1500$ 
the diversity is still very high and the algorithm does not exploit 
good solutions because the selective pressure on the population is too low.

Figure \ref{divers-anisotropy} represents the local diversity along generations
for a cGA with anisotropic selection. Values of $\alpha$ increase from 
top to bottom. By monitoring the anisotropy degree, 
 we can influence on the dynamics of propagation of good solutions.
For low values of $\alpha$, good solutions roughly propagate in circles as 
for a cGA using binary tournament selection.  
When $\alpha$ reaches values close to $1$, the good solutions tend 
to colonize the columns of the grid. The 
diversity is conserved between the columns, which indicate that the algorithm
converge toward different solutions in each columns. Thus, the anisotropic 
selection favors the formation of subpopulations in the columns of the grid \cite{acri06}.
 Crossovers between 
subpopulations then allow the algorithm to explore the search space, as 
long as the probability of selecting participants from different columns 
for the tournament is 
not too low (i.e. $\alpha$ is not too high).
 When $\alpha$ is too high, the selective pressure on the population is too low and 
negatively affects performance.

\section{Discussion}

In this section we summarize and discuss the results on takeover time, performance and 
genotypic diversity and we compare the cGAs using anisotropic selection 
and stochastic tournament selection. 
Two cGAs using different selection operators have been tested 
on instances of QAP. The two selection operators allow to control 
the selective pressure on the population. The analysis of takeover time 
and genotypic diversity 
show the influence of the two operators on the selective pressure. 

When looking at the performance on QAP, we can see that on each instance and for 
both methods, the performance increases as the selective pressure drops 
down until a threshold value of the control parameter. After this value, 
the performance decreases as the selective pressure continue to drop down. 
The threshold values of the control parameter stand in the same range 
on all instances for both of the methods. 
Nevertheless, we notice from table 1 that the takeover times are not similar for 
$\alpha_o$ and $r_o$. Consequently,
 the selective pressure induced on the population is different for the 
two algorithms. The observations on figure \ref{opt-divglob} are in adequation 
with this: The algorithm with stochastic tournament preserves more genotypic diversity 
for the threshold value than the one with 
anisotropic selection. The genotypic diversity measures were made on instance nug30 for 
which the cGA with stochastic tournament selection obtains the best performance. 
However, results on diversity put in evidence properties of the selection operators 
which are independant from the instance tested.

The selective pressure is related to the exploration/exploitation trade-off. 
We conclude from the results presented in table 1 and figure \ref{divglob} 
that studying the exploration/exploitation trade-off is insufficient to 
explain performance of cellular genetic algorithms. In cGAs, the grid topology 
structures the search dynamic.

\onecolumn

\begin{figure}
\begin{center}
\fbox{\begin{tabular}{cccccc}
    &
$1$ &
$300$ &
$500$ & 
$1000$ & 
$1500$ \\

$0.3$ &
\includegraphics[width=2cm,height=2cm]{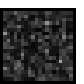} &
\includegraphics[width=2cm,height=2cm]{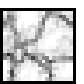} &
\includegraphics[width=2cm,height=2cm]{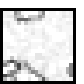} &
\includegraphics[width=2cm,height=2cm]{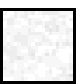} &
\includegraphics[width=2cm,height=2cm]{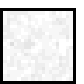} \\
$0.85$ &
\includegraphics[width=2cm,height=2cm]{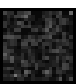} &
\includegraphics[width=2cm,height=2cm]{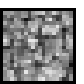} &
\includegraphics[width=2cm,height=2cm]{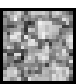} &
\includegraphics[width=2cm,height=2cm]{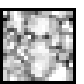} &
\includegraphics[width=2cm,height=2cm]{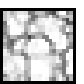} \\
$1$ &
\includegraphics[width=2cm,height=2cm]{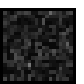} &
\includegraphics[width=2cm,height=2cm]{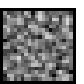} &
\includegraphics[width=2cm,height=2cm]{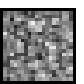} &
\includegraphics[width=2cm,height=2cm]{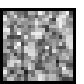} &
\includegraphics[width=2cm,height=2cm]{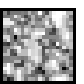} \\
\end{tabular}}
\end{center}
\caption{Local diversity in the population along generations (left to right) for increasing 
$r$ values of stochastic tournament(top to bottom) }
\label{divers-stochastic}
\end{figure}

\begin{figure}
\begin{center}
\fbox{\begin{tabular}{cccccc}
    &
$1$ &
$200$ &
$500$ & 
$1000$ & 
$1500$ \\
$0.4$ &
\includegraphics[width=2cm,height=2cm]{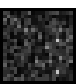} &
\includegraphics[width=2cm,height=2cm]{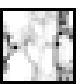} &
\includegraphics[width=2cm,height=2cm]{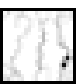} &
\includegraphics[width=2cm,height=2cm]{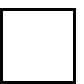} &
\includegraphics[width=2cm,height=2cm]{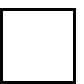} \\
$0.92$ &
\includegraphics[width=2cm,height=2cm]{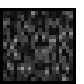} &
\includegraphics[width=2cm,height=2cm]{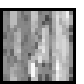} &
\includegraphics[width=2cm,height=2cm]{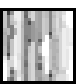} &
\includegraphics[width=2cm,height=2cm]{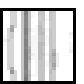} &
\includegraphics[width=2cm,height=2cm]{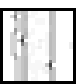} \\
$0.98$ &
\includegraphics[width=2cm,height=2cm]{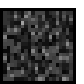} &
\includegraphics[width=2cm,height=2cm]{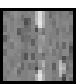} &
\includegraphics[width=2cm,height=2cm]{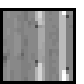} &
\includegraphics[width=2cm,height=2cm]{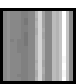} &
\includegraphics[width=2cm,height=2cm]{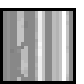} \\
\end{tabular}}
\end{center}
\caption{Local diversity in the population along generations (left to right) for increasing 
$\alpha$ (top to bottom) }
\label{divers-anisotropy}
\end{figure}

\twocolumn

 The overlapped neighborhoods allow to control 
the diffusion of phenotypic and genotypic informations through the population. 

Figures \ref{divers-stochastic} and \ref{divers-anisotropy} show that the 
 algorithm with stochastic tournament or anisotropic selection exploit
differently the structure of the grid. When using the anisotropic selection, 
the algorithm can favor the propagation of solutions vertically. This structuration 
creates subpopulations in the columns of the grid, and solutions can occasionally share 
information with adjacent subpopulations. On the other side, the stochastic 
tournament selection provides an isotropic propagation of solutions. The algorithm 
can control the speed of the propagation by decreasing the probability to select 
the best participant to the tournament as genitor. The snapshots and the figures 
\ref{divglob} and \ref{opt-divglob} show that the genotypic diversity in the population
 is influenced by the exploitation of the grid structure.

The selection operator plays an 
important role in the exploration of the search space and in the exploitation 
of solutions. The two operators we compare allow to control the selective pressure. 
For $r_o$ and $\alpha_o$ the selective pressure induced 
on the population gives the best ratio between exploration and exploitation. 
But this ratio is dependant of the exploration and exploitation dynamics of the 
algorithm. Thus, it is dependant of the selection operator used.
 The measures on genotypic diversity and the snapshots show these 
differences whether the algorithm uses the stochastic tournament or the 
anisotropic selection. Thus, the selective pressure needed to find the best 
exploration/exploitation trade-off is dependant of the transmission mode of 
information through the grid. Furthermore, the existence of a threshold 
value for the parameter which controls the selective pressure do not find explanation 
in the statistic measures on genotypic diversity and takeover time.

A study of the relations between topologic, phenotypic and genotypic distances 
should give a better explanation of performance and as a consequence should 
explain the takeover time and diversity during the search process. 
In order to explain performance of cGAs, we need to study the transmission mode 
of the informations through the grid since the ratio between exploration and exploitation 
seems to rely on it.


\section*{Conclusion and perspectives}

This paper presents a comparative study of two selection operators, the 
anisotropic selection and the stochastic tournament selection, that allow 
a cellular Genetic Algorithm to control the selective pressure on the population.
A study on the influence of the selection operators on the selective pressure 
is made by measuring the takeover time and the genotypic diversity.  
We analyse the average performance obtained on three instances of the 
well-known Quadratic Assignment Problem. A threshold value for the parameters 
of both of the selection operators that gives optimal performance has been 
put in evidence. These threshold values give the adequate selective pressure 
on the population for the QAP. However, the selective pressure is different 
for the two methods. A study on the genotypic diversity shows that the dynamic 
of diffusion of informations through the grid is different when using the stochastic 
tournament or the anisotropic selection operator. The anisotropic selection 
 favors the formation of subpopulations in the columns of the grid, whereas 
the stochastic tournament selection slows down the propagation speed of 
the good solutions. The selection operator have some influence on the dynamic
 of transmission of the information through the grid
and the ratio between exploration and exploitation is not sufficient
to explain the performance of a cGA. 

Nevertheless, we show that even if it is different for the anisotropic selection 
and the stochastic tournament selection, the selective pressure has some influence 
on performances. Further works will analyze the dynamic of diffusion of the 
information through the grid and explain the existence of a threshold value 
for the two cGAs by studying statistic measures on the relations between topologic, genotypic 
and phenotypic distances.

\bibliographystyle{abbrv}


\end{document}